\documentclass{article}
\usepackage{spconf,amsmath,epsfig}

\usepackage[utf8]{inputenc}
\usepackage{spconf}
\usepackage{amssymb}
\usepackage{subfigure}
\usepackage{algorithm}
\usepackage[usenames]{color}
\usepackage{algorithmic}

\title{ LARGE MARGIN FILTERING FOR SIGNAL Sequence Labeling}

\name{Rémi Flamary, Benjamin Labbé, Alain Rakotomamonjy\thanks{
This work is funded in part by the FP7-ICT Programme of the European Community, under the PASCAL2 Network of Excellence, ICT-
216886 and by the French ANR Project ANR-09-EMER-001.
}}
\address{LITIS EA 4108, INSA-Université de Rouen\\76801 Saint-Étienne-du-Rouvray, France\\}

\newcommand{\benjamin}[2]{{\color{red}#2}}
\def\X{\ensuremath{X}}
\def\W{\ensuremath{W}}
\def\F{\ensuremath{F}}
\def\w{\ensuremath{\mathbf{w}}}

\def\Xf{\ensuremath{\widetilde{X}}}

\def\y{\ensuremath{\mathbf{y}}}

\def\dbR{{\mathrm{I\hskip-2.2pt R}}}
\def\dbR{\mathbb{R}}

\def\H{H}

\begin{document}

\maketitle

\begin{abstract}
Signal Sequence Labeling consists in predicting a sequence of labels
given an observed sequence of samples. A naive way is to filter the signal  in order to
reduce the noise and to apply a
classification algorithm on the filtered samples. We propose in this paper to jointly
learn the filter with the classifier leading to a large margin
filtering for classification. This method allows to
learn the optimal cutoff frequency and phase of the
filter that may be different from zero. Two methods are proposed and tested on a toy dataset and on a real life
BCI dataset from \emph{BCI Competition III}.
\end{abstract}

\begin{keywords}
Filtering, SVM ,BCI , Sequence Labeling 
\end{keywords}

\section{Introduction}
\label{sec:introduction}

%The problem we propose to deal with is the one of signal
%sequence labeling. 
The aim of signal sequence labeling is to assign a label to
each sample of a multichannel signal while taking into account the
sequentiality of the samples.
% For instance, a filtering of the signal can be done in order to
% smooth it and to avoid misclassified
% samples. % In this context, our problem is not only to
% find the class changes along the signal but also to know the classes
% between this changes.  This problem can be seen as a structured
% learning problem.
This problem typically arises in speech signal segmentation or in Brain
Computer Interfaces (BCI). Indeed, in real-time
BCI applications, each sample of an electro-encephalography
signal has to be interpreted as a specific command for
a virtual keyboard or a robot hence
the need for sample labeling \cite{bcicometitioniii,millan04}. 

%online
%decoding of mental tasks using EEG continuous signals is often
%needed~\cite{millan04}.
%This consists in
%assigning a label to each sample (or set of samples) of
%the EEG signal.
%For instance each label of the sequence can be interpreted as a
%specific command for the BCI application.  

Many methods and algorithms have already been proposed for signal
sequence labeling.
For instance, Hidden Markov Models~(HMM)~\cite{infmakov} are statistical models
 that are able to learn a joint probability distribution of
  samples in a sequence and their labels. 
% Conditional Random Fields 
%  (CRF)~\cite{Lafferty01Conditional} learn the conditional probability
%  of the labels given the samples (instead of
%  a generative model). 
In some cases, Conditional Random Fields 
(CRF)~\cite{Lafferty01Conditional}
have been shown to outperform the HMM
approach as they do not suppose the observation are independent.
 Structural Support Vector Machines~(Struct-SVM), which are SVMs that learn a mapping from structured
 input to structured output, have also been considered for signal
 segmentation \cite{svmstruct}.  
Signal sequence labeling can also be viewed from a very
 different perspective by considering a
 change detection method coupled with a supervised classifier. For
 instance, a Kernel Change Detection algorithm \cite{desobry03} can be
 used for detecting abrupt changes in a signal and afterwards a classifier
 applied for labeling the segmented regions.

% But due to the unsupervised approach of this method, it is not
% possible to provide a label or a class for the segmented regions.
%For instance to
%provide a good interfacing , a mental task classifier must have a
%quick response.

In order to preprocess the signal, a filtering is often applied and
the resulting filtered samples are used as training
examples for learning. Such an approach poses the issue of
the filter choice, which is oftenly based on prior knowledge
on the information brought by the signals. 
Moreover, measured signals
and extracted features may not be in phase with the labels and a time-lag
due to the acquisition process appears in the signals.  For example, 
in the problem of decoding 
arm movements from brain signals, there exists a natural
time shift between these two entries, hence in their works, 
Pistohl et al.~\cite{Pistohl2008} had to select by a validation method
a delay in their signal processing method.

In this work, we address the problem of automated tuning of
the filtering stage including its time-lag. Indeed, our objective
is to adapt the preprocessing filter and all its properties by including
its setting into the learning process. Our hypothesis
is that by fitting properly the filter to the classification
problem at hand, without
relying on ad-hoc prior-knowledge, we should be able to considerably improve
 the sequence labeling performance.
%These methods work well when the channels and the labels are in phase,
%but do not adapt to time-lag between feature channels and
%labels. Moreover if we suppose that the features are noisy it seems
%sensible to filter the channels. 
So we propose  to take into account
the temporal neighborhood of the current sample directly into the
decision function and the learning process, leading to an automatic
setting of the signal filtering.

For this purpose,  we first propose a naive approach based
on SVMs which consists in considering, instead of a given time
sample, a time-window around the sample.  
% work with the time-windows
%corresponding to the samples  instead of the
%samples alone. 
This method named as Window-SVM, allows us to learn a
spatio-temporal classifier that will adapt itself to the signal time-lag. 
Then, we introduce another approach denoted
as Filter-SVM  which dissociates
the filter and the classifier. 
This novel method jointly learns  a
 SVM classifier and  FIR
filters coefficients. 
By doing so, we
can interpret our filter as a large-margin filter for
the problem at hand. 
 These two methods are
tested on a toy dataset and on a real life BCI signal 
sequence labeling  
problem from \emph{BCI Competition III}~\cite{bcicometitioniii}.%\cite{bcicometitioniii}.

\section{Large margin filter}
\label{sec:method}

\subsection{Problem definition}

%\section{Optimal filtering}
%\label{sec:optimal-filtering-1}

Our concern is a signal sequence labeling problem : we want to obtain a
sequence of labels from a  multichannel time-sample of a signal
or from multi-channel features extracted from
that signal. 
We suppose that the training samples  are gathered in
a matrix 
%we already
%have noisy discriminative features corrupted by noise in a matrix
 $\X\in\dbR^{N\times d}$ containing $d$ channels and $N$ samples.  
%\benjamin{of $d$ channels and $N$ samples.}{from $N$ samples of $d$ signals.}
$\X_{i,j}$~is the value of channel~$j$ for
the $i^{th}$ sample. The vector $\y\in\{-1,1\}^N$ contains the class of each sample.

% \begin{figure}[ht]
%   \centering
%   \includegraphics[width=8cm]{imgs/figmatrixs}
%   \caption{Matrix definition}
%   \label{fig:matdef}
% \end{figure}

% In this paper, we focus on the SVM linear case which leads to this
% minimization problem for the unfiltered features:
% % \begin{equation}
% %   \label{eq:svm}
% %  \min_{w , w_0} \quad  J_{SVM}(w,w_0)  \\ 
% % \end{equation}
% % with:
% \begin{equation}
%   \label{eq:jsvm}
%   J_{SVM}(w,w_0)= \frac{1}{2} {||w||}^2 + \frac{C}{2}
% \sum_{i=1}^{n}\H(\X_{i ,.},\y_i,w,w_0)^2\\ 
% \end{equation}
% where $\H(\X_{i ,.},\y_i,w,w_0)=\max (0 , 1-\y_i(\X_{i ,.}w+w_0))$ is
% the SVM hinge loss. Chapelle\cite{chap_primal} showed that in the
% linear case, the resolution of
% this problem may be solved more efficiently in the primal with a
% gradient descent algorithm. But the standard SVM loss is  not
% differentiable, so a squared version of the hinge loss is used in
% the second part of equation \eqref{eq:jsvm}.

% We propose in this paper to build classifiers taking into account the
% temporal neighborhood of the samples. For that, we propose two different
% approaches. Firstly we work on a windowed version of the samples to
% learn a SVM classifier. Secondly we learn a time filtering of the
% signals jointly with a linear classifier.
In order to reduce noise in the samples
or variability in the features, an usual approach is to filter $\X$ before the classifier learning stage. In literature,
all channels are usually filtered with the same filter (Savisky-Golay
for instance in \cite{Pistohl2008}) although there is no reason
for a single filter to be optimal for all  channels. Let us
define the filters applied to $\X$ by the matrix $\F\in \dbR^{f\times
  d}$. Each column of $\F$ is a filter for the corresponding channel
in $\X$ and $f$ is the size of the FIR %\footnote{Finite Impulse
                                %Response} 
filters. %The defined matrixes may
%be seen on Figure \ref{fig:matdef}.

We define the filtered data matrix $\Xf$ by:
\begin{equation}
  \label{eq:1}
  \Xf_{i,j}=\sum_{m=1}^f ~ \F_{m,j} ~ \X_{i+1-m+n_0,j}
\end{equation}
where the sum is a unidimensional convolution of each channel by the filter in
the appropriate column of $F$. $n_0$ is the delay of the filter, for
instance $n_0=0$ corresponds to a causal filter and  $n_0=f/2$
corresponds to a filter centered on the current sample.

%%% Local Variables: 
%%% mode: latex
%%% TeX-master: "flamary"
%%% End: 

\subsection{Windowed-SVM (W-SVM)}
\label{sec:windowed-svm}

As highlighted by Equation (\ref{eq:1}), a filtering stage
essentially consists in taking into account for 
a given time $i$, instead of the sample $\X_{i,\cdot}$, a 
linear combination of its temporal neighborhood.
 %In order to take into account the time neighborhood of the samples, we
%propose to use \benjamin{for classification}{} a temporal window  around
%the current sample in the classification process instead of the sample alone. 
However, instead of introducing a filter $F$, it is possible
to consider for classification a temporal window  around
the current sample. Such an approach 
would lead to this decision function for the $i^{th}$ sample of $\X$:
\begin{equation}
  \label{eq:classwinSVM}
 f_W(i,\X)=\sum_{m=1}^{f}\sum_{j=1}^{d}\W_{m,j}\X_{i+1-m+n_0,j}+w_0
\end{equation}
where $\W\in \dbR^{f\times d}$ and $w_0\in \dbR$ are the
classification parameters and $f$ is the size of the time-window.
Note that $\W$ plays the role of the filter and the weights of a
linear classifier. In a large-margin framework, 
$\W$ and $w_0$ may be learned by minimizing
this functional:
\begin{equation}
  \label{eq:funcwinsvm}
  J_{WSVM}(W)= \frac{1}{2} ||\W||_F^2 + \frac{C}{2}
\sum_{i=1}^{N}\H(\y,\X,f_W,i)^2\\ 
\end{equation}
where ${||\W||}_F^2=\sum_{i,j}\W_{i,j}^2$ is the squared Frobenius norm
of $\W$, $C$
is a regularization term to be tuned and  $\H(\y,\X,f,i)=\max (0, 1-\y_if(i,\X))$ is the SVM hinge loss. By vectorizing
appropriately $\X$ and $\W$, problem (\ref{eq:funcwinsvm})
 may be transformed into a linear SVM.  Hence, we
can take advantage of many linear SVM solvers existing
in the literature such as the one proposed
by   Chapelle~\cite{chap_primal}. 
%Note that Chapelle gives
%the complexity  of the linear SVM
By using that solver, Window-SVM complexity is 
about $\mathcal{O}(N.(f.d)^2)$ which scales quadratically with
the filter dimension.

%In this equation the
%hinge loss is squared , leading to a differentiable cost.%  One of the limits is that
% the complexity of the problem is $f\times d$ and the size of the
% learning vector is $(N,f.d)$ which may not fit in memory for
% long filters \benjamin{}{if this problem is implemented into an optimizing computer program}.

%This method yields a matrix $\W$ which gives the classification
%weight of different channels and time-lag,
%corresponding to a spatio-temporal filtering. 
%\benjamin{}{\large globalement sur la derniere phrase, j'ai rien compris.}

The matrix $\W$ weights  the importance of
each sample value $\X_{i,j}$ into the decision function.
Hence, channels may have different weights
and time-lag. Indeed, $\W$ will automatically adapt to a phase
difference between the sample labels and the channel signals. 
However, in this method since space and time are
treated independently, $\W$  does not take into account the
multi-channel structure and the
sequentiality of the samples. 
Since the samples of a given channel are known  to
be time-dependent due to the underlying physical process, 
it seems preferable to process them with a filter
and to classify the filtered samples. So we propose in the sequel another method that jointly learns  the time-filtering and a linear classifier on the filtered
sample defined by Eq. \eqref{eq:1}.

% it seemed to us that this method does not
% take into account the fact that we are working with signals. So we
% decided to keep the standard SVM formulation defined in Equ. \eqref{eq:jsvm} \benjamin{with}{on}
% $\Xf$ the filtered signals \benjamin{defined}{from} Equ. \eqref{eq:1}.

\subsection{Large margin filtering (Filter-SVM)}
\label{sec:optimal-filtering}

We propose to find the filter $\F$ that maximizes the margin of the linear
classifier for the filtered samples.  
In this case, the decision function is:
\begin{equation}
  \label{eq:decisionsigsvm}
   f_F(i,\X)=\sum_{m=1}^{f}\sum_{j=1}^{d}\w_j\F_{m,j}\X_{i+1-m+n_0,j}+w_0
\end{equation}
where $\w$ and $w_0$ are the parameters of the linear SVM
classifier corresponding to a weighting of the channels.
By dissociating the filter and the decision function
weights, we expect that some useless channels (non-informative or too noisy) 
for the decision function get small weights.
Indeed, due to the double weighting $\w_j$ and $\F_{.,j}$,
and the specific channel weighting role played by 
$\w_j$, this approach, as shown in the experimental
section is able to
perform channel selection.

The decision function given
in Equation (\ref{eq:decisionsigsvm}) can be obtained by minimizing:
\begin{equation}
  \label{eq:svmopt}
J_{FSVM}=\frac{1}{2}||\w||^2+\frac{C}{2}\sum_{i=1}^{n}\H(\y,\X,f_F,i)^2 + \frac{\lambda}{2}
{||\F||}^2_F 
%\frac{1}{2} {||w||}^2_2+ \frac{C}{2}+\sum_{i=1}^{n}\H(\Xf_{i
%,.},\y_i,w,w_0)^2+ 
\end{equation}
w.r.t. $(F,\w,w_0)$ where ${||\F||}_F$ is the Frobenius norm,
 and $\lambda$ is a regularization term to be tuned. Note that without the
 regularization term ${||\F||}^2_F$, the problem is ill-posed. Indeed,
in such a case, one can always decrease $||\w||^2$ while keeping the empirical hinge loss constant by multiplying $w$ by $\alpha<1$ and
 $\F$ by $\frac{1}{\alpha}$ .

% A regularization term has to be added for $\F$ otherwise
% one can always find a better solution by multiplying $w$ by $k<1$  and
% $\F$ by $\frac{1}{k}$.

\benjamin{\large Et la non convexite, on en fait quoi? C'est pas
  annodin non plus, il faut prouver que les multiples minimums sont
  équivalents, ou expliciter que l'ajout de la minimisation de la
  norme sur F permet de limiter le nombre solutions.}{}

The cost
defined in Equation (\ref{eq:svmopt}) is differentiable and provably non-convex when jointly optimized with respect to 
all  parameters. However, $J_{FSVM}$ is differentiable and convex with respect to
$\w$ and $w_0$ when $F$ is fixed as it corresponds to a linear SVM with
squared hinge loss. Hence, for a given value of
$F$, we can define
$$
J(F) = \min_{\w,w_0} \frac{1}{2}||\w||^2+\frac{C}{2}\sum_{i=1}^{n}\H(\y,\X,f_F,i)^2
$$
% $J_{SSVM}$ is convex and
%differentiable with respect
%to $F$ for fixed values of $\w$ and $w_0$. 
which according to Bonnans et al.  \cite{bonnans_pertubation}
is differentiable. Then if $\w^*\text{ and }w_0^*$ are the optimal values
for a given $F^*$,  the gradient of the second term of $J(\cdot)$
% (see Bonnans and Shapiro \cite{bonnans_pertubation}) 
with respect to $F$ at 
the point $F^*$ %,$w^*\text{, }w_0^*$) 
is:
\begin{equation}\label{eq:gradF}\nonumber
  \nabla_{\F_{m,j}} J(F^*)=-\sum_{i =1 }^N\y_i(\w^*_j\X_{i-m+1+n_0,j})\times\H(\y,\X,f_{F^*},i)
 % (1-\y_i(\sum_k w_k\Xf_{i,k}+w_0))
\end{equation}
% But it is interesting to note that minimizing $J_{SSVM}$
% with a fixed $F$ boils down to a SVM problem for which we can use
% existing methods. Moreover if $F^*,w^*\text{ and }w_0^*$ are the
% optimal values Bonnans and Shapiro \cite{bonnans_pertubation} showed that  :
% \begin{equation}
%   \label{eq:minsigsvmsol}
%   \min_{F}J_{SSVM}(F,w^*,w_0^*)=\min_{F,w,w_0}J_{SSVM}(F,w,w_0)
% \end{equation}
Now,  since $J(F)$ is differentiable and since its value 
can be easily computed by a linear SVM, we choose
for learning the decision function to 
minimize  $J(F) + \frac{\lambda}{2} \|F\|_F^2$ 
with respects to $F$ instead  of minimizing problem (\ref{eq:svmopt}).
Note that due to  the 
objective function non-convexity in problem (\ref{eq:svmopt}),
these two minimization problems are not strictly equivalent, but
our approach has the advantage of taking into account
the intrinsic large-margin structure of the problem.
 \begin{algorithm}[ht]
 \caption{ Filter-SVM solver \label{algorithm1}}
 \begin{algorithmic}
 \STATE Set $\F_{l,k}=1/f$ for $k=1\cdots d$ and $l=1\cdots f$
 %$\beta_k= 1$,
%\STATE Compute  $K_k $ kernel matrices for all tasks
%\STATE \comRF{it\'eration du K ? pas cod\'e comme ça...}
\REPEAT
 % \STATE Solve SVM with filter $\F$
  \STATE $D_F \leftarrow$ gradient of $J_{FSVM}$ with respect to $\F$
 % \STATE k$\leftarrow$LineSearch along $D_F$ 
  \STATE $(\F,w^*,w_0^*)\leftarrow$ Line-Search along $D_F$ 
  %\STATE $(w,w_0)\leftarrow$ SVM Solve with $\X$ filtered by $\F$
 % \STATE Update $\F$
%\\STATE $d_k^{t+1} \leftarrow d_k^{t}+ \gamma_t D_{t,k}$  
\UNTIL{ Stopping criterion is reached}
\end{algorithmic}
\end{algorithm}
% The gradient of the second term in Equ.  \ref{eq:svmopt} respect to $\F$ is:
% \begin{equation}
%   \label{eq:gradF}
%   \nabla J_{\F_{m,n}}=-\sum_{i \in I}\y_i(w_m\X_{i-m+1+n_0,n})\times\H(\y,\X,f_F,i)
%  % (1-\y_i(\sum_k w_k\Xf_{i,k}+w_0))
% \end{equation}
% where $I=\{i\in I, 1-y_i(\sum_k w_k\Xf_{i,k}+w_0)>0\}$ contains the indexes
% of the misclassified vectors. For the gradient descent, the gradient of the
% Frobenius norm was added to Equ. \ref{eq:gradF} to take into account
% the regularization.

 For solving the optimization
problem, we propose a gradient descent algorithm along $\F$ with a line
search method for finding the optimal step.
The method is detailed in algorithm~\ref{algorithm1}.
Note that at each computation of $J(F)$ in the line search, the 
optimal $\w^*\text{ and
}w_0^*$ are found by solving a linear SVM.  
The iterations in the algorithm may be stopped by two stopping criteria: a
 threshold on the relative variation of $J(F)$ or a threshold on
 variations of $\F$ norm.

Due to the non-convexity
of the objective function, it is difficult to provide
an exact evaluation of the solution complexity. 
However, we know that the gradient computation  has
order of
$\mathcal{O}(N.f.d)$ and that when $J(F)$ is computed at each  
step of the line search, a $\mathcal{O}(N.d^2)$ linear SVM is
solved and a $\mathcal{O}(N.f.d)$ filtering is applied.

%%% Local Variables: 
%%% mode: latex
%%% TeX-master: "flamary"
%%% End: 

% \begin{equation}
% \label{eq:decFtaskt}
% f_t(x) = \sum_{k=1}^M f_{t,k}(x) + b_t \quad\quad \forall t \in \{1,\cdots, T\} 
% \end{equation}

\section{Results}
\label{sec:results}

\subsection{Toy Example}
\label{sec:toy-example}

We use a toy example that consists of $nbtot$ channels, only $nbrel$ of them being discriminative.
Discriminative channels have a switching mean $\lbrace-1,1\rbrace$
controlled by the label and corrupted by a gaussian noise of deviation $\sigma$.
%Toy channels where generated from two Gaussian distributions of same standard
%deviation $\sigma$ but with different means depending on the
%label~$\lbrace-1,1\rbrace$.
The length of the regions with constant label follows a
uniform distribution law between $[30,40]$~samples and different time-lags are applied to the channels.
% The training set is composed of generated signals of size $1K$~samples,
% the test set is composed of signals of size $3K$~samples.
%We generate $nbtot$ channels, only $nbrel$ of them are discriminative.
 We selected $f=21$  and $n_0=11$ corresponding to a good average
 filtering  centered on the current sample.
\begin{figure}[ht]
  \centering
\includegraphics[width=\columnwidth,height=3.5cm]{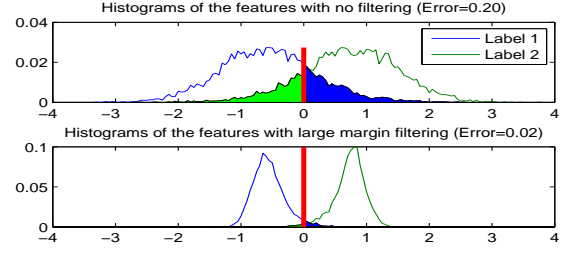}
  \caption{Histograms of both labels with and without filtering
    (vertical axis are different) for a 1 channel signal with $\sigma=1$}
  \label{fig:repartquigere}
\end{figure}
Figure \ref{fig:repartquigere} shows how the
samples are transformed thanks to the filter $F$ for a unidimensional signal.
In this case, the mean test error due to the noise is 16\% for the
unfiltered signal, while only 2\% for the optimally filtered signal.

% \subsubsection{HMM comparison}
% \label{sec:hmm-comparison}
Window-SVM and Filter-SVM are compared to SVM without
filtering, SVM with an average filter of size $f$ (Avg-SVM) and  HMM with
a Viterbi decoding. 
%CRF are not compared as they have the
%same limit as HMM for signals with time-lag. 
The regularization parameters are selected by a
validation method.
The size of the signals is of $1000$~samples for the learning
and the validation sets and of $5000$~samples for the test set.
All the processes are run ten times, the test error is the the average over the runs.

% \begin{figure}[ht]
%   \centering
% \includegraphics[width=\columnwidth,height=5cm]{imgs/acc_noise}
% %\includegraphics[width=6.5cm]{imgs/visuclasse}
%   \caption{Test error for different $\sigma$ values $(nbtot=30,nbrel=3)$}
%   \label{fig:toyvarsigma}
% \end{figure}

\begin{figure}[t]
  \centering
\includegraphics[width=\columnwidth,height=3.5cm]{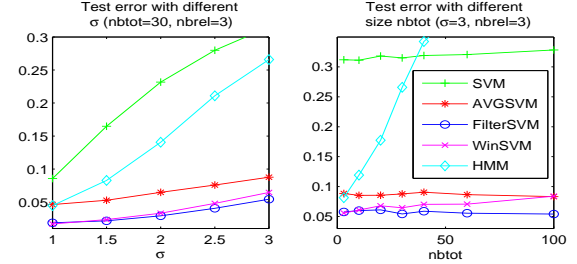}
  \caption{Test error for different $\sigma$ values
    ($nbtot=~30$, $nbrel=~3$, on the left) and for different number of
    channels $nbtot$ ($\sigma=~3$, $nbrel=~3$, on the right) } 
 \label{fig:resultstoy}
\end{figure}
\begin{figure}[ht]
  \centering
\includegraphics[width=\columnwidth,height=2.8cm]{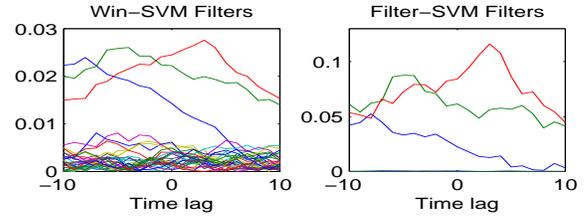}
  \caption{Coefficients of $W$ (left) and coefficients $F$ weighted by
    $w$ (right) for $nbrel=3$, $nbtot=30$, $\sigma=3$}
  \label{fig:compwinsig}
\end{figure}

The methods are compared for different $\sigma$ 
values with ($nbtot=30$, $nbrel=3$). The test error is plotted
on the left of Figure
\ref{fig:resultstoy}. We can see that only Avg-SVM, Window-SVM  and
Filter-SVM adapt to 
time-lags between the channels and the labels. Both Window-SVM and
Filter-SVM outperform the other methods, even if for a heavy noise,
the last one seems to be slightly better.
% \begin{figure}[ht]
% \label{fig:toyvarnbtot}
%   \centering
% \includegraphics[width=\columnwidth,height=5cm]{imgs/acc_size}
% %\includegraphics[width=6.5cm]{imgs/visuclasse}
%   \caption{Test error for different $\sigma$ values $(nbtot=30,nbrel=3)$}  
% \end{figure}
Then we test our methods for a varying number of channels in order to see how
dimension is handled ($nbrel=3$, $\sigma=3$). 
%Test error may be seen in Figure \ref{fig:resultstoy} (right\benjamin{plot}{}).
%First we observe that the HMM methods based on estimation of
%probabilities do not handle properly high dimensional data.
Figure \ref{fig:resultstoy} (right\benjamin{plot}{}) shows the
interest of Filter-SVM over Window-SVM in hight dimension as we can
see that the last one
tends to lose his efficiency, and even to be similar to Avg-SVM. 
This comes from the fact that Filter-SVM can more efficiently perform a
channel selection thanks to the weighting of $\w$. Figure
\ref{fig:compwinsig} shows
 the filters returned by both methods. We observe that only the
coefficients of the relevant signals are important and that the other
signals tend to be eliminated by small weights for Filter-SVM,
explaining the better results in high dimension.

% We compared the test error of classification for different noise
% values. We calculated the test error for a gaussian mixture (MG), a
% SVM without filtering (SVM), SVM with an average filter of size 11,
% HMM and SignalSVM with an online filter of size 11. The regularization
% term $\lambda_2$ has in this case has a small impact on the result and
% was set to 100.

% \begin{table}[ht]
% \label{tab:tabhmmcomp}
%   \centering
%   \begin{tabular}{|c|c|c|c|}
%     \hline 
%     $\sigma$  &1&1.5&2\\ \hline
%  MG&16.0$\pm$.4&25.6$\pm$.7&31.6$\pm$\\
%  SVM&16.2$\pm$.4&26.0$\pm$.8&31.7$\pm$.9\\
%  HMM&3.4$\pm$.7&17.6$\pm$2.1&42.6$\pm$3.5\\
%  SVM(avg)&24.0$\pm$1.8&24.8$\pm$2.1&26.7$\pm$2.2\\
%  SSVM&9.8$\pm$1.5&15.9$\pm$1.8&20.7$\pm$1.9\\ \hline
%   \end{tabular}
%   \caption{Test error for different noise values}
% \end{table}
% Result are shown Table \ref{tab:tabhmmcomp}. We can see that for
% a small noise, HMM perform better than our method. But when the noise
% is important the filtering becomes essential and the Gaussian mixture
% becomes more and more overlapping, which implies probabilities near
%from 1/2 and an increased transition matrix regularization.

% \subsubsection{Average filter comparison}
% \label{sec:aver-filt-comp}

% We compared our filtering to the average filtering in the case of a
% high dimensional signal with different delays for each variable.

\subsection{BCI Dataset}
\label{sec:bci-dataset}

We test our method on the BCI Dataset from \emph{BCI Competition III}~\cite{bcicometitioniii}. The
problem is to obtain a sequence of labels out of brain activity
signals for 3 human subjects. The data consists in 96~channels containing PSD features (3~training sessions, 1~test session,
$N\approx{3000}$ per session) and the
problem has 3~labels (left~arm, right~arm or feet). 

We use Filter-SVM that showed
better result in hight dimension for the toy example. The multi-class aspect
of the problem is handled by using a One-Against-All strategy.
The regularization
parameters are tuned using a grid search validation method on the third
training set. We compare our method to the best BCI competition
results (using only 8 samples) and to the SVM without filtering. Test error for
different filter size $f$ and delay $n_0$ may be seen on Table \ref{tab:bcidataset}.
%------------
\begin{table}[t]
  \centering
\begin{tabular}{|l|c|c|c|c|} \hline
Method & Sub 1 & Sub 2 & Sub3 & Avg\\
\hline
BCI Comp.   & 0.2040 & 0.2969 & 0.4398 & 0.3135 \\ \hline
SVM         & 0.2877 & 0.4283 & 0.5209 & 0.4123 \\ \hline
Filter-SVM  &        &        &        &        \\
$f=8,n_0=0$ & 0.2337 & 0.3589 & 0.4937 & 0.3621 \\
$f=20,n_0=0$& 0.2021 & 0.2693 & 0.4381 & 0.3032 \\
$f=50,n_0=0$& \textbf{0.1321} & \textbf{0.2382} & \textbf{0.4395} &
\textbf{0.2699} \\ \hline
Avg-SVM  &        &        &        &        \\
$f=100,n_0=50$& 0.1544 &0.2235  & 0.3870  & 0.2550 \\ \hline
Filter-SVM  &        &        &        &        \\
$f=100,n_0=50$& \textbf{0.0537} & \textbf{0.1659} & \textbf{0.3859} &
\textbf{0.2018} \\ \hline
\end{tabular}
  \caption{Test Error for BCI Dataset}
  \label{tab:bcidataset}
\end{table}
%--------
%
%---------
\begin{figure}[t]
  \centering
\includegraphics[width=\columnwidth,height=3cm]{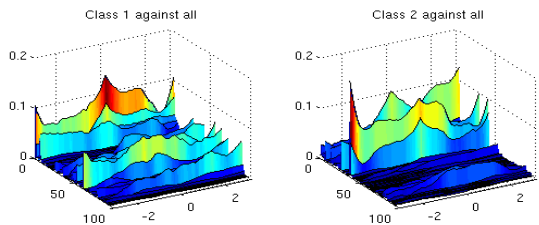}
  \caption{$F$ filters (subject 1) for label 1 against
    all (left) and label 2 against all (right).}
  \label{fig:visuBCI}
\end{figure}
%--------
% \begin{table}[t]
%   \centering
% \begin{tabular}{|l|c|c|c|c|} \hline
% Method & Sub 1 & Sub 2 & Sub3 & Avg\\
% \hline
% BCI Comp.   & 0.2040 & 0.2969 & 0.4398 & 0.3135 \\ \hline
% SVM         & 0.2877 & 0.4283 & 0.5209 & 0.4123 \\ \hline
% $f=8,n_0=0$ & 0.2454 & 0.3638 & 0.4980 & 0.3691 \\
% $f=20,n_0=0$& 0.1835 & 0.3096 & 0.4579 & 0.3170 \\
% $f=50,n_0=0$& \textbf{0.1361} & \textbf{0.2368} & \textbf{0.4189} & \textbf{0.2639} \\ \hline
% $f=100,n_0=50$& 0.0468 & 0.1768 & 0.3589 & \textbf{0.1942} \\ \hline
% \end{tabular}
%   \caption{Test Error for BCI Dataset}
%   \label{tab:bcidataset}
% \end{table}
%
Results show that one can improve drastically the result by using
longer filtering with causal filters ($n_0=0$). Note that
Filter-SVM outperform Avg-SVM with a centered filter. 

Another advantage of this method is that one can visualize a discriminative space-time map (channel selection, shape of the filter and delays).
We show for instance in Figure~\ref{fig:visuBCI} the discriminative filters $F$ obtained  for
subject~1, and we can see that the filtering is extremely different
depending on the task.

The Matlab code corresponding to these results will be provided on our
website for reproducibility. 

%%% Local Variables: 
%%% mode: latex
%%% TeX-master: "flamary"
%%% End: 

\section{Conclusions}
\label{sec:conclusion}

We have proposed two methods for automatically learning a spatio-temporal
filter used for  multi-channel signal classification.  
Both methods  
have been tested on a toy example and on a real life dataset from
\emph{BCI Competition III}. 

Empirical results clearly show the
benefits of adapting the signal filter to the
large-margin classification problem despite the non-convexity of the criterion. 
%We also want to emphasize

% \textbf{Je ne vois pas l'interet de ca puisque un
% Avg-SVM peut aussi etre implemente dans un DSP, la vraie
% difference etant que ton filtre est plus malin}

%Both methods  are tested on a toy example and on a real life dataset from \emph{BCI Competition III} and show the
%interest to learn a large margin filtering of the channels. Note that
%the decision step of
%>>>>>>> .r210
%this classification methods may be efficiently implemented in a
%Digital Signal Processor for a Real-Time application because it is
%based on vector/matrix multiplication.
%\benjamin{}{(Tu parles de l'optimisation ou seulement de la phase de décision?)}

In future work, we plan to extend our approach to  non-linear
case, we believe that a differentiable kernel can be used instead of
inner products at the cost of solving the SVM in the dual
space. Another
perspective would be to adapt our methods to the multi-task situation,
where one wants to  jointly learn one matrix $F$ and several classifiers (one
per task). 

%%% Local Variables: 
%%% mode: latex
%%% TeX-master: "flamary"
%%% End: 

%\section*{Bibliographie}

\small
\bibliographystyle{IEEEbib}
\bibliography{biblioRF,biblioAR}

\begin{thebibliography}{1}

\bibitem{bcicometitioniii}
B.~Blankertz et~al.,
\newblock ``{The BCI competition 2003: progress and perspectives in detection
  and discrimination of EEG single trials},''
\newblock {\em IEEE Transactions on Biomedical Engineering}, vol. 51, no. 6,
  pp. 1044--1051, 2004.

\bibitem{millan04}
J.~del~R Mill\'an,
\newblock ``On the need for on-line learning in brain-computer interfaces,''
\newblock in {\em Proc. Int. Joint Conf. on Neural Networks}, 2004.

\bibitem{infmakov}
O.~Capp\'e, E.~Moulines, and T.~Ryd\`en,
\newblock {\em Inference in Hidden Markov Models},
\newblock Springer, 2005.

\bibitem{Lafferty01Conditional}
J.~Lafferty, A.McCallum, and F.~Pereira,
\newblock ``Conditional random fields: {P}robabilistic models for segmenting
  and labeling sequence data,''
\newblock in {\em Proc. 18th International Conf. on Machine Learning}, 2001,
  pp. 282--289.

\bibitem{svmstruct}
I.~Tsochantaridis, T.~Joachims, T.~Hofmann, and Y.~Altun,
\newblock ``Large margin methods for structured and interdependent output
  variables,''
\newblock in {\em Journal Of Machine Learning Research}. 2005, vol.~6, pp.
  1453--1484, MIT Press.

\bibitem{desobry03}
F.~Desobry, M.~Davy, and C.~Doncarli,
\newblock ``{An online kernel change detection algorithm},''
\newblock {\em IEEE Transactions on Signal Processing}, vol. 53, pp.
  2961--2974, 2005.

\bibitem{Pistohl2008}
T.~Pistohl, T.~Ball, A.~Schulze-Bonhage, A.~Aertsen, and C.~Mehring,
\newblock ``Prediction of arm movement trajectories from ecog-recordings in
  humans,''
\newblock {\em Journal of Neuroscience Methods}, vol. 167, no. 1, pp. 105--114,
  Jan. 2008.

\bibitem{chap_primal}
O.~Chapelle,
\newblock ``Training a support vector machine in the primal,''
\newblock {\em Neural Comput.}, vol. 19, no. 5, pp. 1155--1178, 2007.

\bibitem{bonnans_pertubation}
J.F. Bonnans and A.~Shapiro,
\newblock ``Optimization problems with pertubation : A guided tour,''
\newblock {\em SIAM Review}, vol. 40, no. 2, pp. 202--227, 1998.

\end{thebibliography}
%\bibliography{/home/alain/recherche/biblio/biblioRF,/home/alain/recherche/biblio/biblioAR}

\end{document}